\documentclass[10pt,twocolumn,letterpaper]{article}

\usepackage[pagenumbers]{cvpr} 

\usepackage{graphicx}
\usepackage{amsmath}
\usepackage{amssymb}
\usepackage{booktabs}

\usepackage{xcolor}
\usepackage[pagebackref,breaklinks,colorlinks]{hyperref}

\usepackage[capitalize]{cleveref}
\crefname{section}{Sec.}{Secs.}
\Crefname{section}{Section}{Sections}
\Crefname{table}{Table}{Tables}
\crefname{table}{Tab.}{Tabs.}
\crefname{table}{Table.}{Table}

\def\cvprPaperID{3973} 
\def\confName{CVPR}
\def\confYear{2022}

\usepackage{comment}
\usepackage{algorithm}
\usepackage{xcolor}
\usepackage{algpseudocode}
\usepackage{diagbox}
\usepackage{multirow}
\usepackage{float}
\usepackage{enumitem}
\usepackage{wrapfig}
\usepackage{makecell}
\definecolor{ruisi}{rgb}{1,0,0}
\definecolor{sai}{rgb}{0,0,1}

\title{Improving Differentiable Architecture Search with a Generative Model} 

\author {Ruisi Zhang, Youwei Liang, Sai Ashish Somayajula, Pengtao Xie\\
UC San Diego\\
ruz032@ucsd.edu}
\usepackage{bibentry}

\begin{document}

\maketitle

\begin{abstract}

In differentiable neural architecture search (NAS) algorithms like DARTS, the training set used to update model weight and the validation set used to update model architectures are sampled from the same data distribution. Thus, the uncommon features in the dataset fail to receive enough attention during training. In this paper, instead of introducing more complex NAS algorithms, we explore the idea that adding quality synthesized datasets into training can help the classification model identify its weakness and improve recognition accuracy. We introduce a training strategy called ``Differentiable Architecture Search with a Generative Model(DASGM)." In DASGM, the training set is used to update the classification model weight, while a synthesized dataset is used to train its architecture. The generated images have different distributions from the training set, which can help the classification model learn better features to identify its weakness. We formulate DASGM into a multi-level optimization framework and develop an effective algorithm to solve it. Experiments on CIFAR-10, CIFAR-100, and ImageNet have demonstrated the effectiveness of DASGM. Code will be made available.
\end{abstract}

\section{Introduction}
Neural architecture search (NAS) finds broad applications in  computer vision, such as image classification\cite{cui2019fast,chen2021bn,yang2021towards}, object detection\cite{wang2020fcos,yao2020sm,tan2020efficientdet},  semantic segmentation\cite{liu2019auto,liu2020scam,lin2020graph}, and so on. 
Differentiable  NAS methods~\cite{Zela2020Understanding,chu2020fair} have attracted much research attention due to their computational efficiency. In these methods, a training set used to update  model weights  and a validation set used to update architectures are sampled from a small dataset like CIFAR-100. During training, the image variance is low and some of the uncommon features in the dataset fail to receive enough attention during searching. Previous works like PC-DARTS~\cite{xu2020pcdarts} designs more complex search model architecture on sampled ImageNet to avoid this. However, it relies heavily on the image sampling quality, and the result may be volatile. 


To solve the problems described above, in this paper, we explore the idea that incorporating synthesized datasets into NAS can help the classification model improve its performance. We propose a training strategy called DASGM, which uses the data samples from the training set to update the model weight and the data samples synthesized by the class conditional GAN to update the model architecture. During searching, the class conditional GAN can produce images with different poses and variance, which do not exist in the original training set. We consider these images to have better features. These better features from different image distributions can help the model identify its weakness and improve training accuracy. 

Differentiable search algorithms mainly consist of second-order algorithms like DARTS, Robust DARTS, and first-order algorithms like P-DARTS and PC-DARTS.
We first start with second-order searching algorithms, which consist of four steps: 1) We first update the classification model weight on the training set. 2) Then, we update the class conditional GAN weight on the validation set. 3) Next, we update the classification model architecture on the dataset synthesized by the generator. 4) Finally, we update the discriminator architecture on the validation set. We train the four steps end-to-end. Then, we introduce a reduced framework for first-order search algorithms. Because heuristically optimizing the architecture search loss on discriminator and classification model makes the training very unstable. In the reduced framework, the training steps are similar to the original framework, but we only search the classification model architecture and set the discriminator architecture to default.

\begin{figure}[t]
	\centering
	\includegraphics[width=0.8\columnwidth]{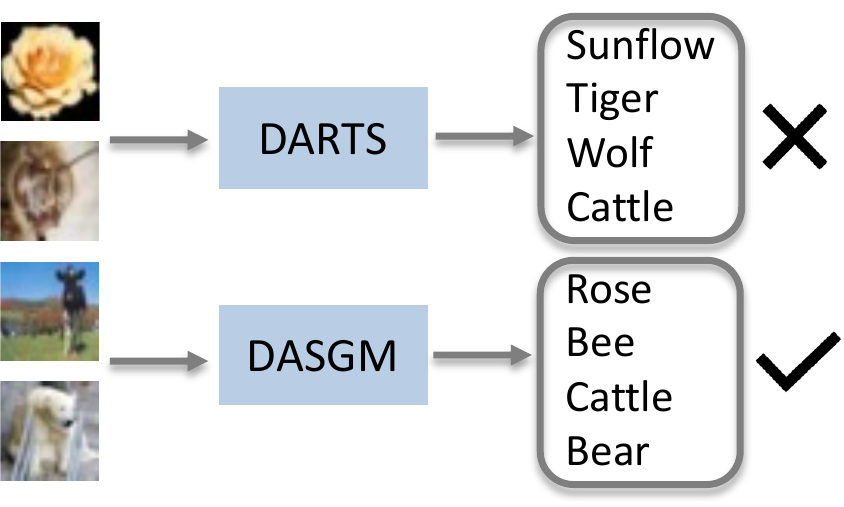}
	\caption{An illustration of how DASGM helps differential NAS algorithms identify their  weakness. The upper result is the prediction from DARTS, and the lower result is the prediction from DASGM+DARTS.
	}  
	\label{fig:overal}
\end{figure}

We intuitively describe our training strategy in Figure~\ref{fig:overal}. The upper prediction is from DARTS, and the lower prediction is from DASGM+DARTS, where we apply our training strategy to DARTS. From the figure, we see that  DASGM+DARTS can make correct prediction on confusing samples like  tiger-like bee and yellow rose, while DARTS cannot. The reason is the class conditional GAN can produce synthesized images with  better features to teach the classification model during searching. Therefore, the classification model can identify its weakness and improve model performance.

The following are the key contributions of our work: 
\begin{itemize}
    \item To the best of our knowledge, our work represents the first one which  generates synthetic validation data for improving  differentiable NAS. Our method significantly improves classification accuracy on CIFAR-100, CIFAR-10, and ImageNet. 
    \item We propose a multi-level optimization framework to formulate DASGM and develop an effective algorithm to solve it.
    \item Our approach can be incorporated into a variety of differentiable NAS algorithms. Experiments on DARTS, P-DARTS, and PC-DARTS have demonstrated the effectiveness of our method.
\end{itemize}

\section{Related Works}
\subsection{Neural Architecture Search}
Neural architecture search (NAS) automatically searches neural architecture with better performance in machine learning tasks. Compared with traditional manually designed models, NAS requires less labor work and achieves better performance. Existing NAS algorithms can be categorized into three groups: 1) Reinforcement learning-based approaches~\cite{zoph2016neural,pham2018efficient,zoph2018learning},  2) Evolution-based approaches ~\cite{liu2017hierarchical,real2019regularized}, and 3) Gradient-based approaches~\cite{cai2018proxylessnas,chu2021darts,xie2018snas}. For reinforcement learning approaches, candidate architectures are sampled from search space based on reinforcement learning algorithms. For evolution-based approaches,  candidate architectures are represented as a population, and poorly-performing architectures are eliminated. For gradient-based approaches, the search space is relaxed to be continuous and optimized based on its validation loss by gradient descent. The first two approaches take expensive computational costs to train a NAS model. Therefore, recent research on NAS mainly focuses on proposing better gradient-based algorithms to sample from search space.

\subsection{Generative Adversarial Networks}
GANs~\cite{DBLP:journals/corr/GoodfellowPMXWOCB14} aim at synthesizing images from random vectors. They use a generator to map random vectors to images, and a discriminator to distinguish generated images from real images. Class conditional Some of the models~\cite{DBLP:journals/corr/MirzaO14,pmlr-v70-odena17a} mainly focus on generating high quality images from class labels in large dataset. There are mainly three approaches to improving image generation quality: 1) designing more complex network architectures for generator and discriminator, 2) introducing better training strategies, and 3) designing more sophisticated loss functions. 

For example, BigGAN~\cite{DBLP:conf/iclr/BrockDS19} applied orthogonal regularization to the generator and can synthesize high-resolution images on large datasets like ImageNet. SAGAN\cite{zhang2019self} introduced a self-attention mechanism into convolutional layers in GANs to help model long-range, multi-level dependencies across image regions. StyleGAN~\cite{DBLP:journals/corr/abs-1812-04948} borrowed network architecture from style transfer networks and enables intuitive, scale-specific control of image synthesis.  

Our paper employs class conditional GANs to generate images as the synthesized dataset to train gradient-based NAS models in the search stage. Previous research like GTN~\cite{pmlr-v119-such20a} also tries to use generative models to accelerate neural architecture search. However, we have two significant differences: 1) GTN mainly accelerates evolution-based NAS approaches, while our work focuses on gradient-based approaches. 2) GTN only updates the generator during training, while our work updates the whole class conditional GAN, including architecture and weight. To the best of our knowledge, our work is the first work to incorporate synthesized datasets in differentiable NAS. The proposed algorithm can be applied to many other differentiable NAS approaches.

\subsection{Multi level Optimization}

Multi-level optimization\cite{sinha2017review} is defined as a mathematical program, where an optimization problem contains another optimization problem as a constraint. It has wide applications in machine learning\cite{bagloee2018hybrid}, supply chain\cite{kuo2011hybrid}, and electricity\cite{ye2019incorporating}. The problem consists of two-level optimization tasks: upper level and lower level. The optimization results are the vectors with lower-level optimal and satisfy upper-level constraints. Our paper solves the optimization problem by giving a descent direction in the upper-level function while keeping the lower-level function at an optimal position.

\section{Problem Formulation}

\begin{figure*}[h]
  \centering
  \includegraphics[width=\linewidth]{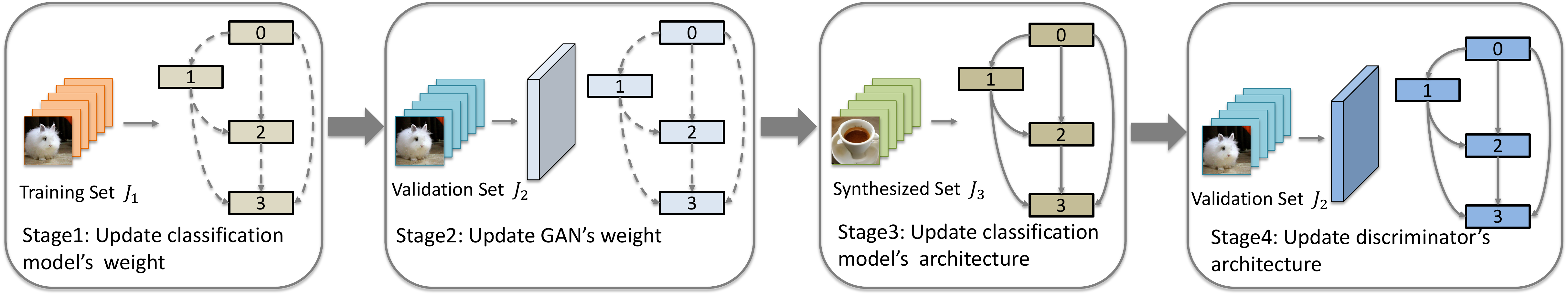}
  \caption{Overview of our training strategy DASGM. Firstly, we update the classification model weight on the training set $J_1$. Then, we update the class conditional GAN weight on the validation set $J_2$. Next, we update the classification model architecture on the dataset synthesized by the generator. Finally, we update the discriminator architecture on the validation set $J_2$.}
    \label{fig:fm}
\end{figure*}

\paragraph{Overview.}  There are two models in DASGM: a classification model as the backbone and a class conditional GAN model as the auxiliary. The class conditional GAN generates new image samples to help the classification model identify its weakness during searching. The two models work together to achieve better classification performance. 

\begin{table}[h]
\centering
\small
\begin{tabular}{l|l}
\hline
Notation & Meaning \\
\hline
$A$ & Architecture of the classification model\\
$W$ & Network weights of the classification model\\
$G$ & Network weights of the generator \\
$B$ & Architecture of the discriminator \\
$E$ & Network weights of the discriminator \\
$J_1$ & Training set I\\
$J_2$ & Validation set II\\
$J_3$ & Images generated by the generator  \\
\hline
\end{tabular}
\caption{Notations used in DASGM}
\label{tb:notations}
\end{table}

As shown in Figure~\ref{fig:fm}, for the second-order search algorithm, DASGM consists of four stages. In the first stage, the classification model trains its network weights $W$ on the training set $J_1$ with the architecture fixed. In the second stage, the class conditional GAN model trains its network weights $G$ and $E$ on the validation set $J_2$ with the architecture fixed. It helps the generator generates image samples with better features. In the third stage, the classification model updates its architecture on the dataset $J_3$ created by the class conditional GAN model. In the fourth stage, the discriminator updates its architecture $B$ on the validation dataset $J_2$. We train the four stages end-to-end. We only use the previous three stages for the first-order search algorithm and set the discriminator architecture to default. The notations are summarized in Table~\ref{tb:notations}.

\paragraph{Objective.} 
We start with the second-order search algorithm because it is more complex compared with first-order search algorithms. 
In the training steps described above, the classification model aims to minimize the classification loss by adjusting its weight during training. The class-conditional GAN model adjusts the generator and discriminator weight to minimize the generator loss while maximising the discriminator loss. We update the classification architecture and the discriminator architecture by minimizing the classification model architecture search loss.

Putting all the objective functions described above together, we get the following optimization framework:

\begin{equation}
    \begin{array}{l}
\max _{B} \min _{A}\;\;  L_{DASGM}\left(A, W^{*}\left(A\right), G^*(B), J_2, J_3\right) \\
s.t. \;\;   
 G^*(B)
      =\textrm{argmin}_{G}\textrm{argmax}_{E} \quad L_{GAN}(B, G, E,J_2)
\\
\quad\;\;\;
W^{*}(A)=\textrm{argmin} _{W}\;\; L_{class}\left(A, W, J_1\right)
\end{array}
\label{eq:learning_objective}
\end{equation}
where $L_{class}\left(A, W, J_1\right)$ is the classification loss, $L_{GAN}(B, G, E,J_2)$ is the class conditional GAN loss and $L_{DASGM}\left(A, W^{*}\left(A\right), G^*(B), J_2, J_3\right)$ is the classification model architecture search loss. $J_1$ and $J_2$ are both sampled from the original training set containing class-image pairs. Given $G^*(B)$, it is used to generate a synthetic validation set $J_3$, which updates the classification model architecture. The objective of our framework is to minimize the classification model architecture search loss  $L_{DASGM}\left(A, W^{*}\left(A\right), G^*(B), J_2, J_3\right)$. It depends on the classification model architecture and the synthesized dataset. For the synthesized dataset, the generated image quality depends on the discriminator architecture. By adjusting the discriminator's architecture, we can have better features generated to help the classification model identify its weakness. 




The optimization formula described above corresponds to the four learning stages in Figure~\ref{fig:fm}. From bottom to up, the first two optimization formulas correspond to the first two stages. The third optimization problem corresponds to the third and fourth stages. The four stages are performed jointly end-to-end in a multi-level optimization framework, where different stages influence each other. The solution $W(A)^{*}$ is obtained by minimizing the training loss $L_{class}\left(A, W, J_1\right)$. The solution $G^*(B)$ is obtained by adjusting $E$ to minimize the training loss $L_{GAN}(B, G, E, J_2)$ and adjusting $G$ to maximize the training loss $L_{GAN}(B, G, E, J_2)$. The obtained $W(A)^{*}$ and $G^*(B)$ are nested on the constraint of the third optimization problem.  The $G^*(B)$ in the second stages is used to generate the augmented dataset $J_3$ to update the classification model architecture in the third optimization problem.

\paragraph{First order search algorithm.}
Differentiable NAS algorithms like DARTS use second-order search algorithms. However, in most recent papers, the first-order search algorithm is applied to speed up training. Applying DASGM to these NAS algorithms makes the model performance significantly improved if we only update the class conditional GAN weight. However, the classification accuracy is reduced if we update the architecture of both the classification model and the discriminator. The reason is we assume $W^{*}(A)$ is the same as $W$ when optimizing the architecture search loss $L_{DASGM}$ in the first-order search algorithm. It will make the search algorithm becomes a heuristic of optimizing the loss on the synthesized dataset.
On the other hand, updating the discriminator architecture requires optimizing the search loss on the validation set simultaneously. As a result, the search can be unstable to fit the parameters on two datasets. Therefore, we also introduce a reduced DASGM consisting of three stages when the first-order search is applied. 


In the first stage, the classification model trains its network weights $W$ on the training set $J_1$ with the architecture fixed. In the second stage, the class conditional GAN model trains its network weights $G$ and $E$ on the validation set $J_2$. In the third stage, the 
classification model updates its architecture $A$ on the synthesized dataset $J_3$. The new optimization framework is summarized below:

\begin{equation}
    \begin{array}{l}
\min _{A}\;\;  L_{DASGM}\left(A, W^{*}\left(A\right),  G^*, J_3\right) \\
s.t. \;\;   
 G^*
      =\textrm{argmin}_{G}\textrm{argmax}_{E} \quad L_{GAN}(G, E,J_2)
\\
\quad\;\;\;
W^{*}\left(A\right)=\textrm{argmin} _{W}\;\; L_{class}\left(A, W, J_1\right)
\end{array}
\label{eq:first_learning_objective}
\end{equation}

\section{Optimization Algorithm}
In this section, we develop a gradient-based optimization algorithm to solve DASGM. Firstly, we update the generator, the discriminator, and the classification model weight in every minibatch. We approximate $W^{*}(A)$ using one-step gradient descent w.r.t $L_{class}\left(A, W, J_1\right)$, where $\eta_{w}$ is the learning rate:

\begin{equation}
\label{eq:learner}
    W^{*}(A)=W'=W-\eta_{W} \nabla_{w} L_{class}\left(A, W, J_1\right). 
\end{equation}

For $G^*(B)$, we approximate it using one-step gradient ascent w.r.t $L_{GAN}(B, G, E, J_2)$, where $\eta_{g}$ is the learning rate:
\begin{equation}
\label{eq:generator}
    G^*(B)=G'=G-\eta_{g}\nabla_{G}L_{GAN}(B, G, E, J_2).
\end{equation}

We update discriminator weight $E$ using one-step gradient descent, where $\eta_{e}$ is the learning rate:
\begin{equation}
\label{eq:discriminator_loss}
    E' = E+\eta_{e}\nabla_{E}L_{GAN}(B, G, E, J_2).
\end{equation}

We plug updated classification model weight $W'$ and updated generator weight $G'$ into objective function $L_{DASGM}\left(A, W^{*}\left(A\right), G^*(B), J_2, J_3\right)$ and get an approximated objective: $L_{DASGM}\left(A, W', G', J_2, J_3\right)$. We update classification model architecture $A$ using gradient descent, where $\eta_a$ is the learning rate:

\begin{equation}
\label{eq:learn_archi}
    A\gets A-\eta_a\nabla_{A}L_{DASGM}\left(A, W', G', J_2, J_3\right),
\end{equation}

In the first-order search algorithm, we skip this stage and change the discriminator architecture to default. For the second-order search algorithm, we also update the discriminator architecture using the following formula. We update discriminator architecture $B$ using gradient ascent,  where $\eta_b$ is the learning rate:
\begin{equation}
\label{eq:dis_archi}
    B\gets B+\eta_b\nabla_{B}L_{DASGM}\left(A, W', G', J_2, J_3\right).
\end{equation}

We summarize the overall algorithm in Algorithm~\ref{alg:opt}. 

\begin{algorithm}[h]
\caption{Optimization algorithm for LPT with a generator}
\label{alg:opt}
\begin{algorithmic}
\While{not converged}
  \State Update classifier weight using Eq.~\eqref{eq:learner}
  \State Update generator weight using Eq.~\eqref{eq:generator}
  \State Update discriminator weight using Eq.~\eqref{eq:discriminator_loss}
    \State Update classifier architecture using Eq.~\eqref{eq:learn_archi}
  \If{second order search}
    \State Update discriminator architecture using Eq.~\eqref{eq:dis_archi}
  \EndIf 
\EndWhile 
\end{algorithmic}
\end{algorithm}

\section{Experiment}

\begin{table*}[h]
    \centering
    \small
    \begin{tabular}{{p{4cm} p{1.6cm} p{1.6cm} p{1.6cm} p{2cm}  p{1.7cm} }}
        Methods & CIFAR-10 Error (\%) & CIFAR100 Error (\%) & CIFAR-10 Param (M) & CIFAR-100 Param (M) & Search Cost (GPU-days)\\
        \hline
        DenseNet\cite{huang2017densely} & 3.46 & 17.18 & 25.6 & 25.6 & - \\
        \hline
        PNAS\cite{liu2018progressive} & 3.41±0.09 & 19.53 & 3.2 & 3.2 & 225 \\
        ENAS\cite{pham2018efficient} & 2.89 & 19.43 & 4.6 & 4.6 & 0.5 \\
        AmoebaNet\cite{real2019regularized} & 2.55±0.05 & 18.93 & 2.8 & 3.1 & 3150 \\
        Hireachical Evolution\cite{liu2017hierarchical} & 3.75 & - & 15.7 & - & 300 \\
        GDAS\cite{dong2019searching} & 2.93 & 18.38 & 3.4 & 3.4 & 0.2 \\
        DropNAS\cite{hong2020dropnas} & 2.58±0.14 & 16.39 & 4.1 & 4.4 & 0.7 \\
        ProxylessNAS\cite{cai2018proxylessnas} & 2.08 & - & 5.7 & - & 4.0\\ 
         \hline
        DASGM+DARTS & \textbf{2.70±0.08} & \textbf{17.14±0.10} & 3.1 & 3.2 & 5.2 \\
        DASGM+DARTS+30\% Val & 2.85±0.08 & 18.49±0.27 & 4.5 & 4.6 & 4.0 \\
        DASGM+DARTS+50\% Val & 2.75±0.02& 18.40±0.27 & 2.8 & 2.8 & 4.0 \\
        DASGM+DARTS+70\% Val & 3.28±0.05 & 18.79±0.21 & 2.3 & 2.4 &  4.0\\
        DARTS-1st\cite{liu2018darts} & 3.00±0.14 & 20.52±0.31 & 3.3 & 3.3 & 1.5  \\
        DARTS-2nd\cite{liu2018darts}  & 2.76±0.09 & 20.58±0.44 & 3.3 & 3.3 & 4.0 \\
        \hline
        DASGM+PDARTS & 2.56±0.04 &  \textbf{16.82±0.03} & 3.4 & 4.1 & 0.5\\
        DASGM+PDARTS+30\% Val & 3.01±0.06 & 18.98±0.16 & 3.0 & 2.2 & 0.3\\
        DASGM+PDARTS+50\% Val & 2.94±0.03 & 17.28±0.23 & 3.0 & 2.9 & 0.3\\
        DASGM+PDARTS+70\% Val & 3.03±0.05 & 18.74±0.13 & 4.2 & 2.6 & 0.3\\
        P-DARTS\cite{chen2019progressive}          & \textbf{2.50} & 17.49 & 3.4 & 3.6 & 0.3\\
        \hline
        DASGM+PCDARTS & \textbf{2.52±0.04} & \textbf{17.42±0.11} & 3.8 & 4.4 & 0.4\\
        DASGM+PCDARTS+30\% Val & 3.43±0.12 & 18.19±0.31 & 2.8 & 2.7 & 0.1\\
        DASGM+PCDARTS+50\% Val & 3.13±0.05 & 17.90±0.12 & 2.5& 2.8 & 0.1\\
        DASGM+PCDARTS+70\% Val & 2.74±0.08 &  18.16±0.29 & 3.3& 2.9 & 0.1\\
        PC-DARTS\cite{xu2020pcdarts}        & 2.57±0.07 & 17.96±0.15 & 3.6 & 3.9 & 0.1\\
        \end{tabular}
    \caption{Comparison with different NAS algorithms on CIFAR dataset. ``DASGM-DARTS+30\% Val" means the validation set portion in $J_3$ is 30\%. In the dataset $J_3$, 70\% images are generated by the pre-trained generator, and the rest 30\% are randomly selected from the validation set. The generator and discriminator are not updated during searching when applying DASGM into DARTS in ``DASGM-DARTS+30\% Val". Others follow the same pattern. We run the experiments above with four different seeds and calculate the mean and standard deviation.}
    \label{tb:cifar-result}
\end{table*}
In this section, we apply DASGM to image classification tasks. We incorporate it into DARTS, P-DARTS, and PC-DARTS and conduct experiments on CIFAR-10, CIFAR-100, and ImageNet to demonstrate the effectiveness of DASGM.

\subsection{Datasets}
We evaluate the performance of DARTS\cite{liu2018darts}, P-DARTS\cite{chen2019progressive}, and PC-DARTS\cite{xu2020pcdarts} on three datasets: CIFAR-10, CIFAR-100, and ImageNet\cite{deng2009imagenet}.  

For DARTS\cite{liu2018darts}, P-DARTS\cite{chen2019progressive} and PC-DARTS\cite{xu2020pcdarts}, we first search the architecture on CIFAR-10 and CIFAR-100. And then, evaluate the cells on CIFAR-10, CIFAR-100, and ImageNet. The CIFAR-10 dataset contains 50K training images and 10K testing images (classified into ten classes equally). We split the original 50K training set into a 25000 new training set and a 25000 new validation set in the architecture search stage. The new training set is denoted as $J_1$, and the new validation set is denoted as $J_2$. In the architecture evaluation stage, we use the original training set to train the model stacking multiple copies of the searched cell and the test set to evaluate model performance.
The CIFAR-100 dataset contains 50K training images and 10K testing images (classified into 100 classes equally). Moreover, the dataset division used in architecture search and evaluation is similar to CIFAR-10. 
The ImageNet dataset contains a training set of 1.3M images and a validation set of 50K images(classified into 1000 classes). We used the model consisting of multiple copies of searched cells on CIFAR-10/CIFAR-100 to evaluate the model performance on ImageNet. The model is first trained on the 1.3M training set and then evaluated on the 50K validation set.

\begin{table*}[htb]
    \centering
    \small
    \begin{tabular}{{p{4.1cm} p{2.7cm} p{2.7cm} p{1.5cm}  p{3.1cm} }}
        Methods & Top-1 Error (\%) & Top-5 Error (\%) & Param (M) & Search Cost (GPU-days)\\
        \hline
        Inception-v1\cite{googlenet} & 30.2 & 10.1 & 6.6 & -\\
        MobileNet\cite{howard2017mobilenets} & 29.4 & 10.5  & 4.2 & -\\
        ShuffleNet 2× (v1)\cite{zhang2018shufflenet} &26.4 & 10.2 & 5 & -\\
        ShuffleNet 2× (v2)\cite{ma2018shufflenet} & 25.1& - & 5 & -\\
        \hline
        NASNet-A\cite{zoph2018learning} & 26.0 & 8.4 & 5.3 & 1800\\
        NASNet-B \cite{zoph2018learning} & 27.2 & 8.7 & 5.3 & 488\\
        NASNet-C\cite{zoph2018learning} & 27.5 & 9.0 & 4.9 & 558\\
        AmoebaNet\cite{real2019regularized} & 24.3 & 7.6 & 6.4 & 3150\\
        PNAS\cite{liu2018progressive} & 25.8 & 8.1 & 5.1 & 225\\
        MnasNet-92\cite{tan2019mnasnet} & 25.2 & 8.0 & 4.4 & -\\
        ProxylessNAS\cite{cai2018proxylessnas}  & 24.9 & 7.5 & 7.1 & 8.3 \\
        SNAS\cite{xie2018snas} & 27.3 & 9.2 & 4.3 & 1.5 \\
        \hline
        DASGM+DARTS-CIFAR10 & \textbf{25.6} & \textbf{8.2} & 4.6 &  5.2\\
        DARTS-2nd\cite{liu2018darts} & 26.7 & 8.7 & 4.7 &4\\
        \hline
        DASGM+PDARTS-CIFAR10 & \textbf{24.5} & \textbf{6.8} & 5.7 &  0.5\\
        P-DARTS-CIFAR10\cite{chen2019progressive} & 24.4 & 7.4 & 4.9 & 0.3\\
        P-DARTS-CIFAR100\cite{chen2019progressive} & 24.7 & 7.5 & 5.1 &0.3\\
        \hline
        DASGM+PCDARTS-CIFAR10 & \textbf{23.9} & \textbf{7.1} & 5.3  & 0.4\\
        PC-DARTS-CIFAR10\cite{xu2020pcdarts}  & 25.1 & 7.8 & 5.3 &  0.1\\
        PC-DARTS-ImageNet\cite{xu2020pcdarts} & 24.2 & 7.3 & 5.3 &  3.8 \\
        \end{tabular}
    \caption{Comparison with different NAS algorithms on full ImageNet dataset.}
    \label{tb:imagenet-result}
\end{table*}

\subsection{Settings}
DASGM can be generalized into any differentiable NAS algorithms. Specifically, we apply it to the following NAS methods: 1) DARTS\cite{liu2018darts} 2)  P-DARTS\cite{chen2019progressive} and 3) PC-DARTS\cite{xu2020pcdarts}. The search space in these methods are similar. The candidate operations include: 3 $\times$ 3 and 5 $\times$ 5 separable convolutions, 3 $\times$ 3 and 5 $\times$ 5 dilated separable convolutions, 3 $\times$ 3 max pooling, 3 $\times$ 3 average pooling, identity, and zero.

For CIFAR-10 and CIFAR-100, the classification model's network is a stack of 8 cells in the architecture search stage, with the initial channel number set to 16. The search is performed for 50 epochs, with a batch size of 64. Other hyperparameters like learning weight and decay rate are set in the same way as DARTS and P-DARTS. 
During the architecture evaluation stage, 20 copies of the searched cell are stacked to form a classification model, with the initial channel number set to 36. Then, we train the network for 600 epochs with a batch size of 96 (for both CIFAR-10 and CIFAR-100).
The search stage is performed on a single Nvidia 3090, and the evaluation stage is performed on a single Nvidia 1080Ti. We also take the architecture searched on CIFAR-10 and evaluate it on the entire ImageNet dataset. We first stack 14 cells (searched on CIFAR-10) to form an extensive network and set the initial channel number as 48. Then, we train the network for 250 epochs with a batch size of 512 on 8 Nvidia 1080Ti.

We use BigGAN\cite{DBLP:conf/iclr/BrockDS19} as the main building blocks to generate the synthesized dataset for model training. The GAN can be replaced by any other class conditional GANs. We use BigGAN to synthesis CIFAR-10 and CIFAR-100 images in the architecture search stage for DARTS, P-DARTS, and PC-DARTS. We apply share embedding, skip connections, and orthogonal regularization into class conditional GAN training. Other hyperparameter settings like learning weight and decay rate follow the default setting in BigGAN.

In ablation study settings, we summarize the results for ``effectiveness of end-to-end training" in Table~\ref{tb:cifar-result}. ``DASGM+DARTS+30\% Val" means the generator and discriminator are not updated during training when applying DASGM into DARTS. In the dataset $J_3$, 70\% images are generated by a pre-trained generator, and the rest 30\% are randomly selected from the validation set. Others follow the same pattern. ``DASGM+DARTS" methods in Table~\ref{tb:cifar-result} are the results of applying DASGM into DARTS. 

\subsection{Results}
Table~\ref{tb:cifar-result} shows the classification error (\%), number of weight parameters (millions), and search cost (GPU days) of different NAS methods on CIFAR datasets. Table~\ref{tb:imagenet-result} shows the classification error (\%), number of weight parameters (millions), and search cost (GPU days) of different NAS methods on entire ImageNet datasets. Compared with baseline models, DASGM achieves better classification accuracy.

We make the following observations from the tables: 1) Our training strategy achieves much lower error rates on CIFAR-10, CIFAR-100, and ImageNet than baseline models. For example, we reached the 17.14±0.10 error rate on DASGM+DARTS CIFAR-100, which has a much better performance than DARTS-2nd. The reason is the classification model can continuously identify its weakness by learning from the new features generated by the class conditional GAN in the searching stage.
2) Compared with baseline models, our searched architecture achieves much lower standard deviations. For example, compared with PC-DARTS on CIFAR-10, the standard deviation achieved by DASGM+PCDARTS is 0.04, while the baseline model is 0.07. It reflects that our searched model architecture is more robust and stable. The reason is that we use both the training dataset and the synthesized dataset for training. Hence, the synthesized dataset can produce better features to help the classification model achieve better performance.

Besides, DARTS and previous methods failed to search architectures on CIFAR-100. For example, for DARTS and most previous NAS search algorithms, the error rate on CIFAR-100 is about 19\%, which is lower than human-designed models like DenseNet. P-DARTS and PC-DARTS achieved better classification accuracy than human-designed models by designing more complex searching algorithms. We find that by simply adding quality synthesized data into training, DASGM+DARTS can achieve higher accuracy on CIFAR-100.

\subsection{Ablation Study}
\paragraph{Effectiveness of end-to-end training} 
To evaluate the effectiveness of end-to-end training, we compare the results with and without class conditional GAN in the searching stage. In these experiments, we use pre-trained generators to generate images and combine them with randomly selected validation samples to form a new $J_2$. We make the following observations: 1) Compared with separate training methods, DASGM achieves higher classification accuracy. The reason is that the generator can update its weight during end-to-end training and generates more diverse examples to train the classification model. 2) Compared with different validation set portions, we get the highest training accuracy when the portion is set to 50\%. 
The reason is some of the generated samples might be hard to recognize and, thus, make the classification model's performance worse. In contrast, the samples from the validation set follow similar distribution as the training set, and some of the images might be very similar. It makes image recognition easier on the validation set. Therefore, we need to trade-off the difficulty and get the best performance when the validation portion is set to 50\%.

\paragraph{Effectiveness of class conditional GAN}
In this section, we evaluate how different types of class conditional GAN can help improve model performance. We change the class conditional GAN to different types and evaluate the classification model performance on CIFAR-100 with DASGM+DARTS.
We use BigGAN\cite{DBLP:conf/iclr/BrockDS19}, StyleGAN2-ADA\cite{karras2020training}, and ContraGAN\cite{kang2020contragan} as class conditional GAN and train them end-to-end. We show the results in Table~\ref{tb:diffGAN}. From the tables, we can conclude that the more complex the class conditional GAN is, the less error rate DASGM+DARTS achieves on CIFAR-100. However, the improvement is marginal. For example, the performance of ContraGAN is better than BigGAN because it added contrastive learning into GAN training. However, the error rate is decreased from 16.98 to 16.93, which is not significant. Because in DASGM, our major focus is generating images with different distributions instead of image generation quality.

\begin{figure*}[htb]
  \begin{center}
    \subfloat[DASGM+DARTS]{\label{fig:archi:ours}\includegraphics[width=0.25\linewidth]{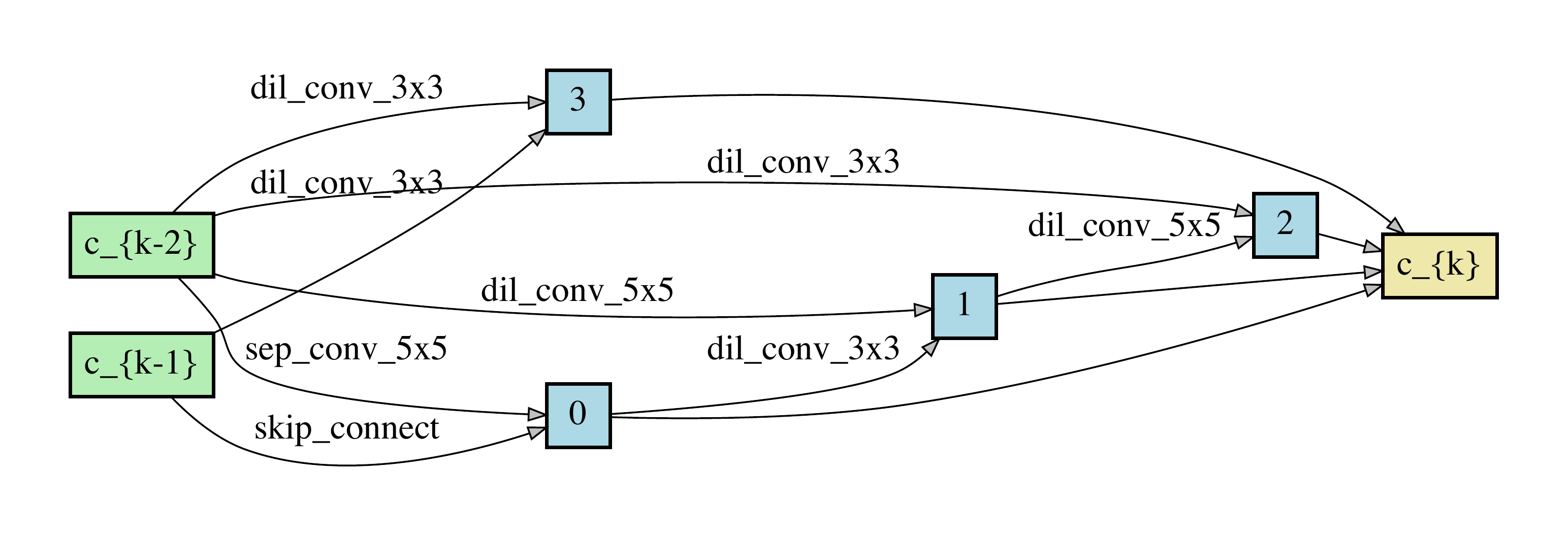}}
    \subfloat[DASGM+DARTS+50\% Val]{\label{fig:archi:pretrain}\includegraphics[width=0.25\linewidth]{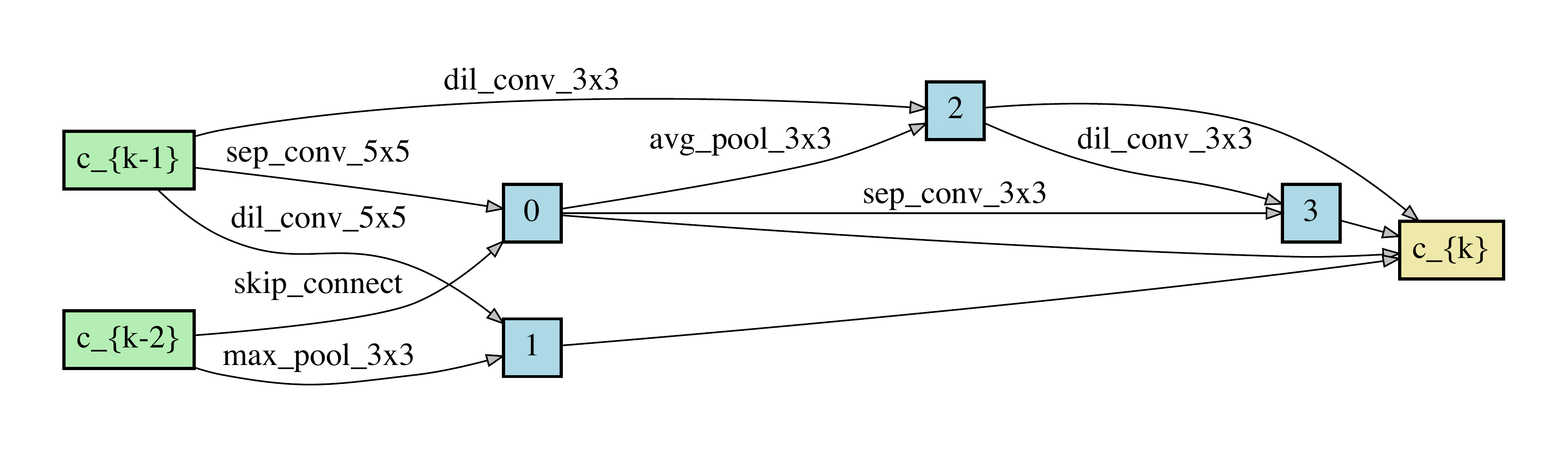}}
    \subfloat[DARTS-1st]{\label{f:archi:darts:1st}\includegraphics[width=0.25\linewidth]{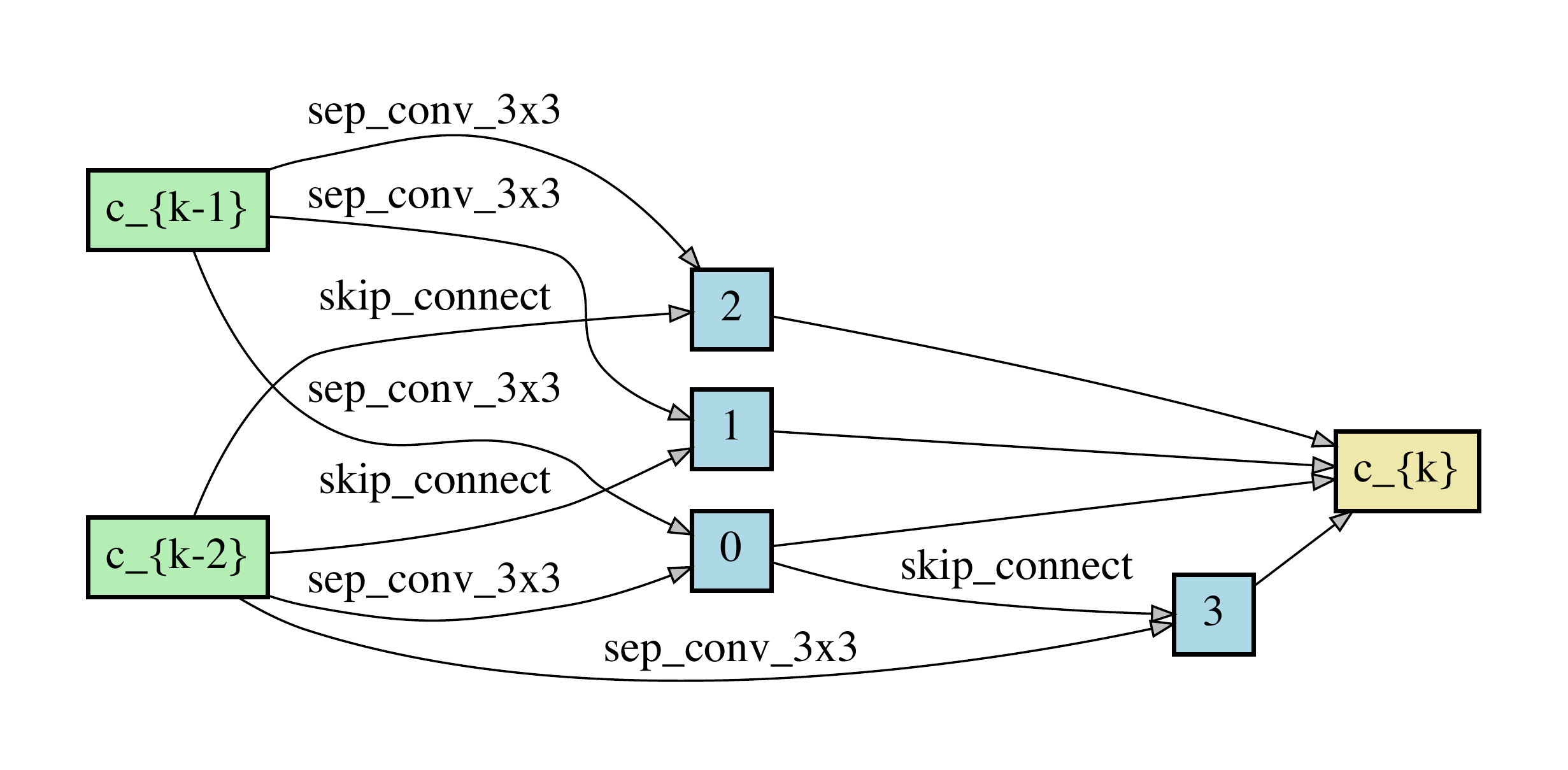}}
    \subfloat[DARTS-2nd]{\label{f:archi:darts:2nd}\includegraphics[width=0.25\linewidth]{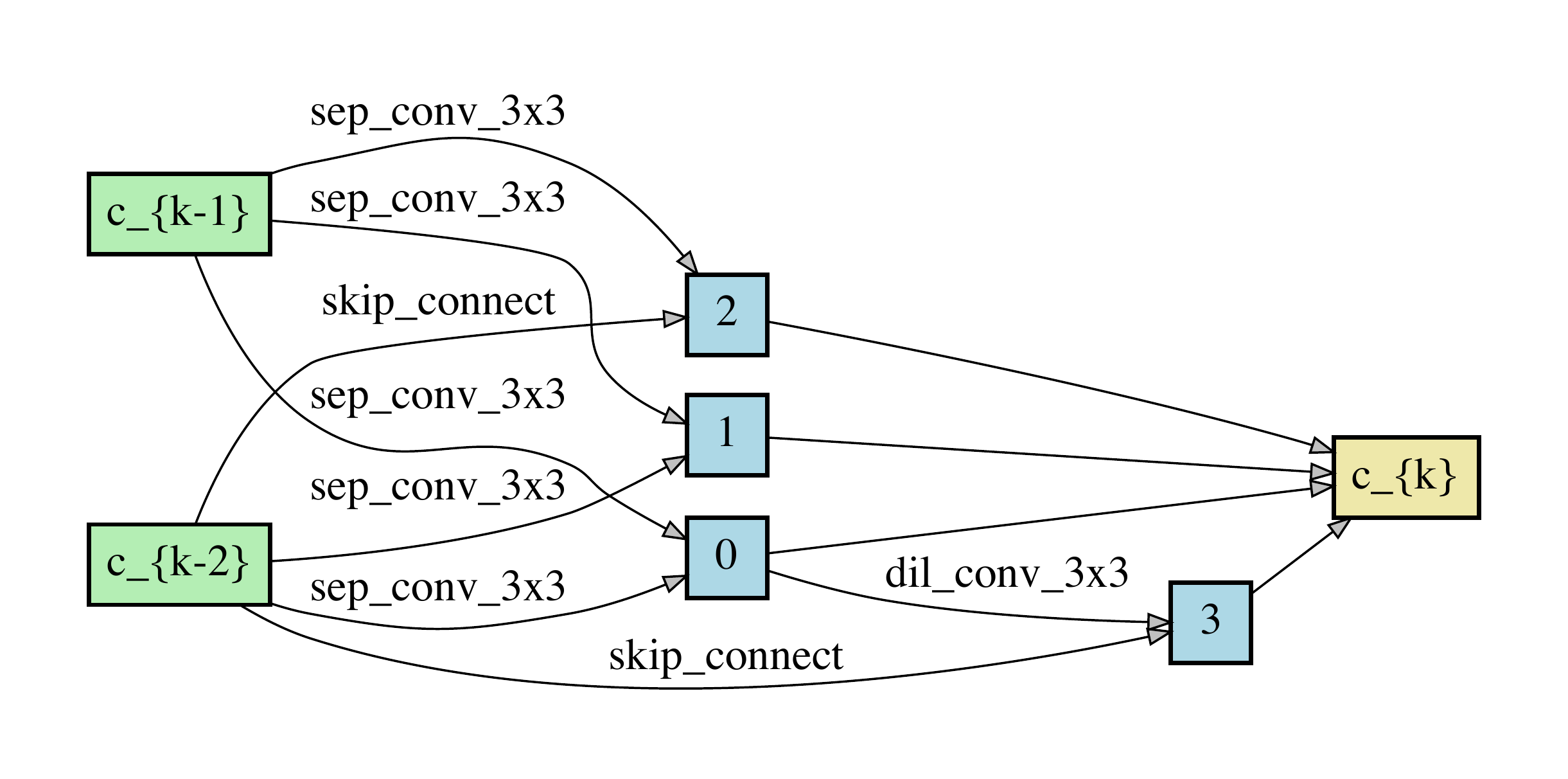}}
  \end{center}
  \caption{Discovered normal cells on DARTS CIFAR-100. From (a)-(d), the algorithms are our approach, with pre-trained GAN, DARTS first order, and DARTS second order.}
  \label{fig:architecture}
\end{figure*}

\subsection{Analysis}
\paragraph{Image Distribution} 
In order to evaluate how generated image distribution changes during the searching stage, we measure the synthesized image distribution using FID and IS scores. The experiment is conducted on CIFAR-100 using DASGM+DARTS. The FID score measures the distribution similarity between generated images and the training set, and the IS score measures the diversity of the generated images. 
The FID and IS score of generated images in every ten epochs are shown in Table~\ref{tb:is}. We can make the following observations from the table: 1) As the generator achieves a higher IS score, the synthesized images are getting more diverse and better quality. 2) We also find that the FID score is around 62, which means the distribution between the synthesized dataset and the training set is very different. Therefore, the synthesized dataset can generate better features to help the classification model identify its weakness.

\begin{table}[htbp]
    \centering
    \small
    \begin{tabular}{{p{3cm} p{3cm}  }}
        Methods & Error Rate(\%)\\
        \hline
        BigGAN &  16.98 \\
        StyleGAN2-ADA &  16.90 \\
        ContraGAN & 16.93  \\
        \end{tabular}
    \caption{Comparison with different class conditional GAN composed into DASGM+DARTS in one run.}
    \label{tb:diffGAN}
\end{table}

\begin{table}[htbp]
    \centering
    \small
    \begin{tabular}{{p{0.7cm} p{1cm}  p{1cm} p{1cm}  p{1cm} p{1cm} }}
        Epoch & 10 & 20 & 30 & 40 & 50\\
        \hline
        FID$\downarrow$ & 62.69 & 63.27 & 63.86 & 64.08 & 62.34\\
        IS$\uparrow$ &  2.39 & 2.38 & 2.41 & 2.42 & 2.44 \\
        \end{tabular}
    \caption{IS and FID score on DASGM+DARTS CIFAR-100 every ten epoch}
    \label{tb:is}
\end{table}

\paragraph{Significance Test}
To determine whether our result is significantly better than baseline models, we also introduce the double-sided t-test in Table~\ref{tb:p-value}. As shown in the table, the p-value of DASGM on CIFAR-10 is higher than 0.05, which means the improvement is not as significant. However, as the classification task becomes complex, DASGM achieves much more significant improvement on CIFAR-100. The reasons are: 1) CIFAR-10 dataset is relatively small with fewer classes than CIFAR-100 and ImageNet. DASGM and baseline models both achieve high classification accuracy on CIFAR-10. Therefore, the improvement is insignificant. 2) As the dataset becomes more complex, by adding image samples with better features to help the classification model identify its weakness, DASGM achieves remarkable improvement on CIFAR-100.

\begin{table}[htbp]
    \centering
    \small
    \begin{tabular}{{p{2cm} p{1.5cm} p{2cm} }}
        Methods & CIFAR-10 & CIFAR-100\\
        \hline
        DARTS-2nd &  0.3575 & 0.0000 \\
        P-DARTS &  0.2722 & 0.0003 \\
        PC-DARTS & 0.2472 & 0.0005 \\
        \end{tabular}
    \caption{p-value of DASGM compared with baseline models.}
    \label{tb:p-value}
\end{table}

\subsection{Visualization}

\paragraph{Search Results}
The results of searched architecture are shown in Figure~\ref{fig:architecture}. From the figure, we display the architecture discovered on CIFAR-10 by DASGM+DARTS, DASGM+DARTS+50\%Val, DARTS first order, and DARTS second order from (a) to (d). By comparing different architectures, DASGM can cascade more layers than the baselines, making the architecture more complex and achieving better classification accuracy.

\begin{figure}[htb]
	\centering
	\includegraphics[width=0.9\columnwidth]{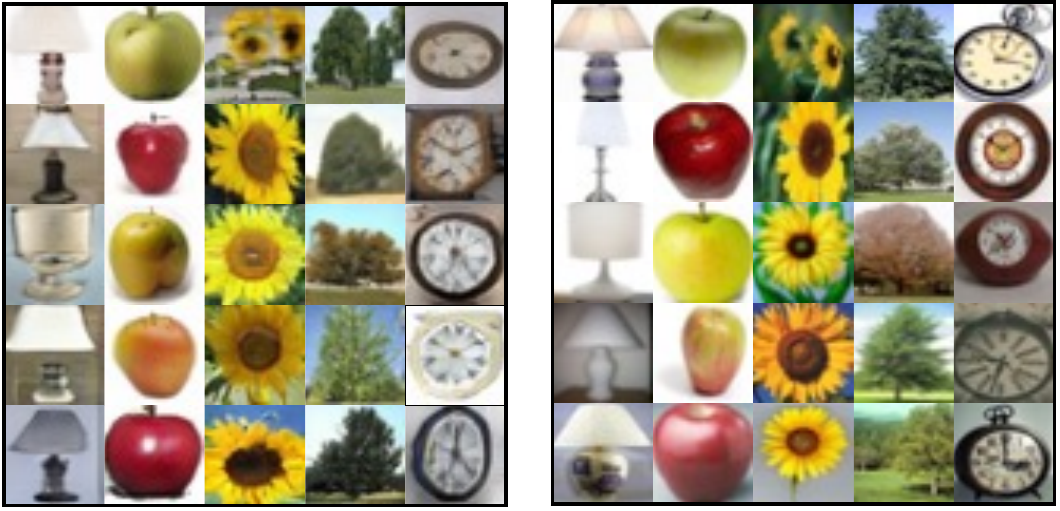}
	\caption{(Left) From top to bottom, we show the generated images in every ten epoch.  (Right) We show that the images belong to the same class from the training set.
	}  
	\label{fig:gen}
\end{figure}

\paragraph{Generated Images}
We show the generated image in the left part of Figure~\ref{fig:gen} and show the images most similar to the synthesized images from the training set in the right part. For the generated images, from top to bottom, we show the images in every ten epochs during the searching stage on CIFAR-100 using DASGM+DARTS. From Figure~\ref{fig:gen}, we make the following observations: 1) From top to bottom, we observe that the generated image quality becomes better after 20 - 30 epochs' training. As training progresses, the synthesized images become more vivid and have more details. 2) Compared with the images on the right half, in which we choose the most similar images from the training set, we find the class conditional GAN model can synthesize images with more poses and variance.

\begin{figure}[htb]
	\centering
	\includegraphics[width=0.9\columnwidth]{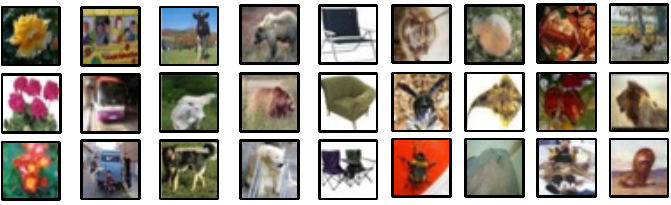}
	\caption{Ssamples from CIFAR-100 test set where DARTS-2nd make the wrong prediction and DASGM+DARTS make correct prediction. The classes are rose, bus, cattle, bear, chair, bee, ray, lobster, and lion from left to right.
	}  
	\label{fig:classify}
\end{figure}

\paragraph{Identify Classification Model Weakness}
We show some of the classification results from the CIFAR-100 test set in Figure~\ref{fig:classify}. The table shows that DASGM is more robust when the image samples have less uncommon features and can correctly classify them. For example, when the flower color becomes yellow in the rose class, DASGM+DARTS can still identify it as a rose. However, for DART-2nd, it is recognized as a sunflower. Therefore, by adding class conditional GAN into searching, our classification model can continuously identify its weakness by learning better features in the synthesized dataset.

\section{Conclusion and Future Work}
We propose a new machine learning training strategy called DASGM, which can continuously improve the performance of differentiable NAS algorithms by adding quality synthesized dataset into searching. Furthermore, we propose a multi-level optimization framework to formalize the problem and develop an effective algorithm to solve it. By continuously identifying model weakness during searching, DASGM has demonstrated its effectiveness on CIFAR-100, CIFAR-10, and ImageNet. 

When designing DASGM, we only focus on general image classification tasks. We will extend DASGM to more high-level applications like semantic segmentation and object detection in the future.

{\small
\bibliographystyle{ieee_fullname}
\bibliography{refs-2}
}
\end{document}